\documentclass[11pt]{article}

\usepackage{acl}
\usepackage{times}
\usepackage{latexsym}
\usepackage[T1]{fontenc}
\usepackage[utf8]{inputenc}
\usepackage{microtype}
\usepackage{inconsolata}
\usepackage{graphicx}
\usepackage{booktabs}
\usepackage{multirow}
\usepackage{amsmath}
\usepackage{array}
\usepackage{siunitx}
\usepackage{algorithm}
\usepackage{algorithmic}
\title{Bounded Path Context: A Controlled Study of Visible Path History in LLM-Based Knowledge Graph Question Answering\thanks{Code: \url{https://github.com/AndyShan11/Bounded-Path-Context}}}

\author{
  Xihang Shan \\
  School of Mathematical Sciences \\
  Xiamen University \\
  \texttt{19020232202354@stu.xmu.edu.cn} \And
  Ye Luo\thanks{Corresponding author.} \\
  School of Informatics \\
  Xiamen University \\
  \texttt{luoye@xmu.edu.cn}
}

\begin{document}
\maketitle

\begin{abstract}
LLM-based knowledge-graph question answering (KGQA) delegates graph traversal to language models, turning each question into a sequence of local relation-selection decisions repeated across beams and hops.
A common but untested default is to serialize the complete partial path into every routing prompt, even though the controller already maintains this path as exact symbolic state.
Bounded Path Context (\textbf{BPC}) decouples these two roles: the controller retains full paths in symbolic memory for answer extraction and audit, while the relation-selection prompt exposes only the question, the current entity, outgoing relation candidates, and at most the last $K$ hops.
A controlled sweep over $K$---fixing graph neighborhoods, beam budget, depth, decoding, and answer-extraction format---shows that bounded histories match or exceed full-history prompting on complete WebQSP and CWQ test sets with Qwen3.5-9B-AWQ\@: $K{=}1$ achieves 0.487 answer-set F1 on WebQSP versus 0.472 for full history, and $K{=}0$ reaches 0.287 on CWQ versus 0.274, with 9.7\% and 12.1\% fewer input tokens respectively.
At the 4B scale, $K{=}1$ remains the strongest setting on both benchmarks.
Per-example analysis reveals that 71--84\% of examples are unaffected by history length, while the affected cases expose when prior hops disambiguate versus distract.
These results suggest that path serialization length is better treated as a tunable interface variable than as a default assumption in LLM-based graph controllers.
\end{abstract}

\section{Introduction}

LLM-based knowledge-graph question answering (KGQA) turns a single question into many small routing decisions over graph neighborhoods.
At each step, a controller asks a language model which outgoing relation to follow next, and the quality and cost of the system depend on this prompt being both informative and cheap to repeat.
Yet a common design choice makes every routing call reread the complete partial path, even though the same path is already stored exactly by the graph controller.

This default matters because iterative KGQA is not a one-shot generation problem.
A single example may require dozens of relation-selection calls as beams expand through entities and relations toward candidate answers.
Serializing the full path into every prompt therefore turns controller memory into model input again and again.
The extra text can consume KV-cache memory, limit local batching, increase latency, and compete with the current relation candidates for attention.
If the complete path is necessary for accurate routing, this cost is justified.
If it is not, then full-history prompting may add repeated input cost and can become a poor default for some graph-agent settings.

This habit persists for a natural reason.
Recent LLM-based graph controllers often make intermediate traces visible to the model, following the broader success of chain-of-thought and tool-use prompting.
Despite these successes, KGQA has a special structure that is easy to overlook: the partial path is a symbolic state maintained by deterministic code, not merely a textual rationale.
The controller can keep entity identifiers, relation identifiers, beam membership, and final evidence paths without asking the LLM to reread them at every hop.
What remains unresolved is whether the complete symbolic path must also be part of the \emph{visible} state for the local routing decision.
Prior systems conflate two roles of context: the path serves as exact bookkeeping for the controller and as optional disambiguating text for the language model.

We study that gap through Bounded Path Context (\textbf{BPC}), a simple prompt-interface intervention for LLM-based KGQA\@.
BPC separates the symbolic search state, which the controller retains in full, from the path history that is shown to the model during relation selection.
At hop $t$, a BPC controller with bound $K$ presents the question, the current entity, the outgoing relation candidates, and only the last $K$ traversed hops.
When $K=0$, the routing prompt contains no previous hop.
When $K=\mathrm{full}$, the prompt recovers the conventional full-history interface.
This intervention does not change the graph neighborhoods or the configured beam and depth budgets.
The controller still stores every retained path exactly; after search, a fixed answer-extraction prompt renders retained complete paths up to the shared prompt budget used in all $K$ settings.

BPC rests on a distinction between control context and explanation context.
For final answers, a complete path can be useful because it supports auditability and evidence inspection.
For the next routing action, however, the question, current entity, and outgoing relation inventory already encode strong local signals.
Older hops can help when the same entity is reached through semantically different routes, but they can also introduce stale entity names, earlier relation labels, and surface forms that are no longer relevant at the current frontier.
BPC therefore asks a narrow empirical question: how much of the accumulated path does the model need for relation selection alone, given that the controller preserves the full symbolic state?

We evaluate this question on the full test sets of WebQSP \citep{yih2016webquestionssp} and ComplexWebQuestions \citep{talmor2018complexwebquestions} with Qwen3.5-9B-AWQ served locally by vLLM \citep{qwen2025qwen25}.
The controlled sweep keeps the graph neighborhoods, beam width, depth, relation cap, decoding temperature, and answer-extraction format fixed while varying the visible path history in relation-selection prompts.
Small visible histories have favorable point estimates relative to the full-history default in this local setting.
On WebQSP, $K\!=\!1$ gives the best answer-set F1 (0.487), $K\!=\!0$ is essentially tied (0.485), and full history obtains 0.472.
On CWQ, $K\!=\!0$ is best (0.287 F1), while full history obtains 0.274 and uses more input tokens.
The gain is not merely a consequence of graph access, since random-relation controls score far below the learned controller.

The result also has an important nuance at the 4B scale.
With Qwen3.5-4B-AWQ under the same controller, full-history prompting has higher F1 than $K\!=\!0$ on both benchmarks, indicating that the smaller model benefits from path context.
However, $K\!=\!1$ has higher F1 than both $K\!=\!0$ and full history at 4B, making one visible hop the most consistently competitive setting across the model sizes studied.
The full path is not the highest-F1 setting at either scale in these runs; the difference is that the 9B model is often strong with zero history, while the 4B model appears to benefit from one hop of local disambiguation.
This boundary is useful for system design because it turns a default prompt convention into a measurable variable.

The paper makes five concrete contributions.
Section~\ref{sec:bpc} defines bounded path context as an interface that decouples exact symbolic state from model-visible path history during relation selection.
Section~\ref{sec:main-results}, Table~\ref{tab:qwen35-main}, and Figure~\ref{fig:k-sweep} provide full-test evidence on WebQSP and CWQ that bounded histories are competitive with full-history prompting for Qwen3.5-9B-AWQ, with up to a 12.1\% reduction in input tokens on CWQ\@.
Section~\ref{sec:controls} adds diagnostic controls, showing that the controlled setting still depends on graph grounding and learned relation selection.
Section~\ref{sec:model-size} and the appendix manifest document a model-size interaction: $K\!=\!1$ has the strongest 4B point estimates and remains competitive at 9B, while full history is not the highest-F1 setting at either scale.
Section~\ref{sec:error-analysis} provides a per-example error analysis showing that most examples have unchanged final F1 under bounded histories, while the changed examples reveal when prior hops help or distract.

\section{Related Work}

\paragraph{Classical KGQA and graph retrieval.}
Knowledge-graph question answering has traditionally been framed as semantic parsing, query-graph search, or learned traversal over structured evidence rather than as free-form text retrieval.
Early semantic-parsing and search-based systems map questions to executable graph queries or ranked query graphs, with WebQSP and CWQ becoming standard testbeds for evaluating whether systems can recover answers from Freebase-style structure \citep{berant2013semantic,yih2016webquestionssp,talmor2018complexwebquestions,lan2020query}.
Neural KGQA systems then add differentiable retrieval, graph propagation, and embedding-based matching, as seen in GraftNet, PullNet, EmbedKGQA, neural symbolic machines, TransferNet, subgraph retrieval, QA-GNN, and unified retrieval-reasoning models \citep{sun2018graftnet,sun2019pullnet,saxena2020embedkgqa,he2021nsm,shi2021transfernet,zhang2022subgraph,yasunaga2021qagnn,jiang2023unikgqa}.
These methods establish the broader premise of this paper: graph structure is not merely background knowledge, but an operational search space whose relations and entities must be selected under limited supervision and limited compute.
Most classical systems, however, do not study how much of an already visited path should be rendered as natural-language context for an LLM at each routing step.
In contrast, BPC keeps the graph, beam budget, and symbolic state fixed, and isolates a narrower interface question: whether complete path histories should be visible to the language model during local relation selection.

\paragraph{LLM-based KGQA and graph controllers.}
Recent work integrates LLMs with KGs through structured interfaces, few-shot logical-form construction, generate-then-retrieve pipelines, discriminative grounding, and joint answer-form decoding \citep{jiang2023structgpt,li2023kbbinder,luo2024chatkbqa,gu2023pangu,yu2023decaf}.
The most direct predecessors to BPC are LLM graph controllers that make graph traversal part of the reasoning process, including Reasoning-on-Graphs, Think-on-Graph, GNN-RAG, and newer agent or retrieval frameworks such as DARA, KELDaR, SubgraphRAG, and graph-constrained reasoning \citep{luo2024rog,sun2024think,mavromatis2024gnnrag,fang2024dara,li2024keldar,li2025subgraphrag,luo2024gcr}.
These systems improve KGQA by decomposing questions, retrieving subgraphs, constraining decoding, or using LLMs to choose actions over graph neighborhoods.
They also make intermediate graph state explicit because the model often needs evidence to plan, verify, or explain its next step.
Yet this design typically treats the visible prompt as the controller's working memory, so the accumulated path is repeatedly serialized even when the controller already stores it exactly.
In contrast, BPC is orthogonal to stronger retrievers, decomposition modules, and constrained decoders: it asks whether the same controller can preserve complete symbolic paths outside the prompt while showing the LLM only a bounded suffix during relation selection.

\paragraph{Reasoning traces, long context, and prompt compression.}
Chain-of-thought and ReAct-style prompting show that intermediate traces and actions can improve language-model reasoning when the model must carry latent state through a task \citep{wei2022chain,yao2023react}.
At the same time, long-context studies show that simply adding more text is not a free improvement, since models can underuse or misread information depending on position and salience \citep{liu2024lost}.
Prompt-compression methods such as Selective Context, LongLLMLingua, RECOMP, and LLMLingua-2 reduce input length by selecting or rewriting textual context, and recent surveys organize these methods into prompt-centric and model-centric compression families \citep{li2023selectivecontext,jiang2023longllmlingua,xu2024recomp,pan2024llmlingua2,li2025promptcompression}.
This line of work is relevant because BPC also reduces model-visible context and treats prompt length as a systems variable that affects latency, memory, and attention.
The difference is that BPC does not compress a document, summarize a rationale, or learn a token selector.
Instead, it exploits a property specific to graph controllers: the full path can remain exact in symbolic memory while only the routing-relevant suffix is exposed to the LLM\@.

\paragraph{Path-based structured representations.}
Path-based code models show that paths can be useful neural inputs when the underlying object is structured \citep{alon2019code2vec,alon2019code2seq,allamanis2018learning}.
We use this work only as an analogy: BPC does not learn path embeddings, but measures which suffix of an exact KG path should be exposed to an LLM controller.

\section{Bounded Path Context}
\label{sec:bpc}

\begin{table}[t]
\centering
\caption{Notation for the BPC controller. The table separates symbolic state variables from prompt-interface variables, which is the distinction used throughout the method.}
\label{tab:notation}
\scriptsize
\setlength{\tabcolsep}{4pt}
\begin{tabular}{p{.22\linewidth}p{.68\linewidth}}
\toprule
Symbol & Meaning \\
\midrule
$q$ & Input question. \\
$G_q=(V_q,E_q)$ & Question-specific directed labeled graph, where $E_q \subseteq V_q \times \mathcal{R} \times V_q$. \\
$S_q$ & Topic entities supplied with $q$. \\
$p_t$ & Symbolic path of length $t$, $(e_0,r_1,e_1,\ldots,r_t,e_t)$. \\
$\tau(p_t)$ & Tail entity $e_t$ of path $p_t$. \\
$D$ & Maximum search depth. \\
$w$ & Per-beam relation width and per-relation tail cap. \\
$B$ & Maximum number of retained beams. \\
$c$ & Maximum number of displayed outgoing relation labels. \\
$K$ & Visible-history bound; $K=0$ hides previous hops and $K=\mathrm{full}$ shows the complete path. \\
$h_K(p_t)$ & Rendered path history exposed to the LLM. \\
$\mathcal{C}_c(e)$ & Capped outgoing relation set shown for entity $e$. \\
\bottomrule
\end{tabular}
\end{table}

Table~\ref{tab:notation} summarizes the notation.
For each question $q$, the controller receives a question-specific directed labeled graph $G_q=(V_q,E_q)$ and a set of topic entities $S_q$.
The implementation builds an adjacency index over triples $(h,r,t) \in E_q$, so outgoing relation lookup and neighbor expansion operate on the local graph rather than by issuing graph queries during inference.
A beam item at hop $t$ is an exact symbolic path
\[
p_t = (e_0, r_1, e_1, \ldots, r_t, e_t),
\]
where $e_0 \in S_q$ and $\tau(p_t)=e_t$ is the current frontier entity.
The complete path is stored as structured triples in controller memory throughout search.
BPC changes only the text rendered to the LLM when selecting the next relation.

Let $\operatorname{suffix}_K(p_t)$ denote the subpath containing the last $K$ hops of $p_t$.
The visible-history function is
\[
h_K(p_t) =
\begin{cases}
\emptyset, & K = 0,\\
\operatorname{suffix}_K(p_t), & 0<K<t,\\
p_t, & K \geq t \text{ or } K=\mathrm{full}.
\end{cases}
\]
In the prompt, the empty case is rendered as an explicit no-history marker rather than by omitting the field, which keeps the prompt format stable across $K$ values.
The controller also renders a deterministic relation inventory
\[
\mathcal{C}_c(e) = \mathrm{first}_c\left(\mathrm{sort}\{r : (e,r,e') \in E_q\}\right),
\]
where $c=50$ in the reported runs.
This cap is a serving and readability budget, not a claim that all outgoing relations are visible to the model: relations beyond the cap remain in the graph index but cannot be selected by the LLM at that routing step.
The LLM receives $q$, $h_K(p_t)$, $\tau(p_t)$, and $\mathcal{C}_c(\tau(p_t))$, and returns at most $w$ relation names or the token \texttt{STOP}.

\begin{algorithm}[t]
\small
\caption{Bounded Path Context controller. The procedure keeps complete symbolic paths in controller memory while bounding only the path suffix shown in relation-selection prompts.}
\label{alg:bpc}
\begin{algorithmic}[1]
\REQUIRE Question $q$, topic entities $S_q$, local graph $G_q$, bound $K$, depth $D$, width $w$, beam budget $B$, relation cap $c$
\ENSURE Predicted answer set $\hat{A}$ and retained symbolic paths $\mathcal{P}$
\STATE Initialize $\mathcal{P} \leftarrow \{(e): e \in S_q\}$ with all paths marked open
\FOR{$t=0$ to $D-1$}
  \STATE Let $\mathcal{A}$ be the open paths in $\mathcal{P}$
  \IF{$\mathcal{A}=\emptyset$}
    \STATE \textbf{break}
  \ENDIF
  \STATE Batch one relation-selection prompt for each $p \in \mathcal{A}$ with nonempty $\mathcal{C}_c(\tau(p))$
  \STATE Parse each LLM response into $\hat{R}(p) \subseteq \mathcal{C}_c(\tau(p))$ with $|\hat{R}(p)| \leq w$
  \STATE Initialize $\mathcal{P}'$ with the finished paths from $\mathcal{P}$
  \FOR{each $p \in \mathcal{A}$}
    \IF{$\mathcal{C}_c(\tau(p))=\emptyset$ or the response is \texttt{STOP} or $\hat{R}(p)=\emptyset$}
      \STATE Mark $p$ finished and add it to $\mathcal{P}'$
    \ELSE
      \FOR{each $r \in \hat{R}(p)$}
        \FOR{each $e' \in \mathrm{first}_w\{v:(\tau(p),r,v)\in E_q\}$}
          \STATE Add $p \circ (r,e')$ to $\mathcal{P}'$
        \ENDFOR
      \ENDFOR
    \ENDIF
  \ENDFOR
  \STATE Retain at most $B$ paths from $\mathcal{P}'$, preferring open paths before finished paths
  \STATE $\mathcal{P}\leftarrow\mathcal{P}'$
\ENDFOR
\STATE Render up to a fixed budget of retained complete paths in $\mathcal{P}$ and ask the LLM to extract comma-separated answer entities
\RETURN $\hat{A}, \mathcal{P}$
\end{algorithmic}
\end{algorithm}

Algorithm~\ref{alg:bpc} summarizes the implemented controller.
Relation selection is batched across active beams, but batching does not change the decision rule.
Responses are parsed as relation names, with permissive exact or substring matching against the displayed candidate set to make the interface robust to minor formatting variation.
If a beam has no outgoing relation, if the model returns \texttt{STOP}, or if no valid displayed relation can be parsed, the beam is marked finished rather than expanded.
For each selected relation, the controller expands up to $w$ tail entities, and the combined beam set is then clipped to $B$ retained paths.
Because the implementation does not learn a beam scorer, this clipping is a deterministic budget rule that preserves open paths first and then finished paths as capacity allows.
This choice keeps the ablation focused on model-visible path history rather than on a second ranking model.

After the search loop, answer extraction is deliberately separated from relation selection.
The extractor receives the first eight retained complete paths, not $h_K(p_t)$.
This prompt budget is shared across $K$ settings, but retained beam order and path sets can differ because relation-selection decisions differ.
Thus $K$ directly changes only routing prompts, while final answer changes can be mediated by which complete paths survive into extraction.

\paragraph{Complexity.}
Let $n=|E_q|$ be the number of triples in the local graph.
Building the adjacency index takes $O(n)$ time and $O(n)$ space per example.
At each hop, at most $B$ open paths issue relation-selection prompts, and each prompt displays at most $c$ relation labels and at most $K$ visible hops when $K$ is finite.
Ignoring LLM internal attention cost, graph-side expansion is bounded by $O(D B (c + w^2))$, since each beam scans a capped relation set and expands at most $w$ relations with at most $w$ tails per relation.
The controller stores at most $B$ paths of length at most $D$, so symbolic state requires $O(BD)$ space after indexing.
The prompt-side benefit of BPC appears in the token term: full history makes visible path text grow as $O(t)$ per routing prompt, while finite $K$ bounds it by $O(\min(K,t))$.
Across a depth-$D$ run with comparable beam survival, full-history routing contributes $O(BD^2)$ path-text units, whereas fixed $K$ contributes $O(BDK)$, which is the mechanism behind the input-token reductions measured in Section~\ref{sec:main-results}.

\paragraph{Design implications.}
BPC is best understood as a separation between deterministic state and model-visible communication.
The controller handles reachability, path storage, beam budgeting, checkpointed execution, and final evidence preservation, while the LLM handles semantic relation selection and answer extraction from retained paths.
This division matters because copying the whole path into every routing prompt treats prompt text as the only memory of the system, even though the code already maintains exact state.
It also reduces a common confound in graph-agent comparisons: the BPC sweep changes $K$ while holding the graph index, relation cap, depth, width, beam budget, deterministic decoding, and answer-extraction format fixed.
The method therefore asks a controlled interface question rather than introducing a new retriever, a new entity-pruning stage, or a new reasoning checker.

\section{Experimental Setup}

\paragraph{Datasets.}
We evaluate on the RoG-aligned local versions of WebQSP and ComplexWebQuestions (CWQ), both of which use Freebase-style graph neighborhoods.
The data loader reads the saved test split from disk, and each example contains an identifier, question, gold answer set, topic entity list, answer entities, and a question-specific graph neighborhood.
Table~\ref{tab:dataset-stats} reports the statistics recoverable from the final result files.
WebQSP has many more gold answers per question on average because list-style questions are common, while CWQ is dominated by single-answer compositional questions.
We do not report average graph size because the final JSONL result files store predictions and search summaries rather than the original triples; the original graph neighborhoods are consumed by the runner when it builds the per-example adjacency index.

\begin{table}[t]
\centering
\caption{Evaluation-set statistics for the RoG-aligned test splits. WebQSP contains many more gold answers per question, while CWQ is mostly single-answer, which motivates reporting answer-set F1 in addition to Hits@1.}
\label{tab:dataset-stats}
\scriptsize
\setlength{\tabcolsep}{4pt}
\begin{tabular}{ll
  S[table-format=4.0]
  S[table-format=2.2]
  S[table-format=2.1]@{\%}}
\toprule
Dataset & Split/source & \multicolumn{1}{c}{$n$} & \multicolumn{1}{c}{Avg.\ gold} & \multicolumn{1}{c}{Single-answer} \\
\midrule
WebQSP & RoG-aligned test & 1628 & 10.20 & 50.0 \\
CWQ & RoG-aligned test & 3531 & 1.89 & 75.8 \\
\bottomrule
\end{tabular}
\end{table}

\paragraph{Models and decoding.}
The main local model is Qwen3.5-9B-AWQ \citep{qwen2025qwen25} served with vLLM \citep{kwon2023efficient}.
We run in language-model-only mode and disable thinking-style generation so that relation-selection outputs remain short and deterministic.
All reported Qwen3.5 runs use temperature 0, seed 42, maximum depth $D=5$, width $w=3$, relation cap $c=50$, and maximum retained beams $B=16$.
The local 9B runs use tensor parallelism 2 on RTX 2080 Ti GPUs, float16 activation compute, AWQ-Marlin quantized weights, eager execution, and text-only vLLM serving.
The maximum model length is 2048 for most runs.
The CWQ full-history configuration requires 4096 because some examples exceed 2048 tokens under complete path serialization; we note this as a potential confound in the Limitations section, since the larger KV-cache allocation can affect vLLM scheduling and throughput.
No model is trained or fine-tuned in this work, so the reported times are inference times for complete evaluation runs.
We keep an API backend in the code path and report small compatibility checks in the appendix, but these checks are not part of the main local comparison and their latency is not mixed with vLLM timing.

\paragraph{Baselines and controls.}
The primary within-controller comparison is the same BPC controller with $K=\mathrm{full}$, which recovers the conventional interface where the full path is serialized into every relation-selection prompt.
This is the cleanest comparison for the interface question because graph neighborhoods, search depth, beam budget, relation cap, decoding, and answer-extraction format are held fixed at configuration time.
The $K=0$, $K=1$, and $K=2$ settings form the bounded-context sweep and serve as hyperparameter ablations over visible history length.
We also include a random-relation control, implemented with the same graph neighborhoods and final answer extractor but with relation and tail choices sampled uniformly rather than selected by the LLM\@.
This control is intentionally not competitive; it tests whether access to the graph alone explains the BPC result.
Finally, Qwen3.5-4B-AWQ rows test whether the visible-history effect changes with model capacity.
We additionally report a chain-of-thought (CoT) diagnostic control that gives the LLM only the question text without graph access, providing a lower bound that isolates the contribution of graph grounding in this setup.
A Think-on-Graph (ToG) reimplementation \citep{sun2024think} adds entity pruning and explicit reasoning checks within the same graph and beam infrastructure; we use it only as a diagnostic comparison against a reasoning-augmented traversal strategy, not as a claim about broad ToG performance.

\paragraph{Checkpointing and audit.}
Full-test local runs are long enough that interrupted jobs are expected on shared machines.
The runner therefore writes one JSONL row per completed example and emits partial summaries.
When restarted, it resumes from existing rows and continues with the next example.
For result reporting, answer metrics are computed from the final complete JSONL files.
For one WebQSP $K=0$ run, cost accounting combines the interrupted first segment and resumed segment; the manifest records this reconstruction explicitly.

\paragraph{Metrics.}
We report Hits@1, answer-set F1 after string normalization, wall-clock time, LLM calls, and estimated input/output tokens.
Hits@1 checks whether the first predicted answer matches any gold answer.
F1 compares normalized predicted and gold answer sets and is therefore more informative for multi-answer questions.
Because all runs use greedy decoding (temperature${=}0$), changing the random seed does not produce meaningful variance.
We therefore report bootstrap 95\% confidence intervals computed by resampling the per-example F1 values 10{,}000 times with replacement; these appear as error bars in Figure~\ref{fig:k-sweep}.
Wall-clock time is useful for end-to-end reproducibility but is sensitive to vLLM scheduling and run interruption, so we treat input tokens and call counts as cleaner efficiency diagnostics when comparing prompt interfaces.
All main BPC runs in Table~\ref{tab:qwen35-main} have zero per-example errors.

\section{Main Results}
\label{sec:main-results}

\begin{table*}[t]
\centering
\caption{Full-test Qwen3.5-9B-AWQ results for the visible-history sweep. Bold marks the best answer metrics within each dataset and underlining marks the second-best values; bounded histories have favorable F1 point estimates relative to full-history prompting in these runs while reducing input-token load. The WebQSP $K=0$ cost is reconstructed from its checkpointed first segment and resume segment, while answer metrics are computed over the final 1,628-row JSONL file.}
\label{tab:qwen35-main}
\scriptsize
\setlength{\tabcolsep}{3.5pt}
\begin{tabular}{ll
  S[table-format=1.3]
  S[table-format=1.3]
  S[table-format=+1.3]
  S[table-format=1.2]@{h\hspace{.45em}}
  S[table-format=3.1]@{K\hspace{.45em}}
  S[table-format=2.2]@{M\hspace{.45em}}
  S[table-format=-2.1]@{\%}
  S[table-format=1.0]}
\toprule
Dataset & History & \multicolumn{1}{c}{Hits@1} & \multicolumn{1}{c}{F1} & \multicolumn{1}{c}{$\Delta$F1} & \multicolumn{1}{c}{Time} & \multicolumn{1}{c}{Calls} & \multicolumn{1}{c}{Input toks} & \multicolumn{1}{c}{$\Delta$In} & \multicolumn{1}{c}{Errors} \\
\midrule
\multirow{4}{*}{WebQSP}
& $K=0$ & {\bfseries 0.648} & \underline{0.485} & +0.013 & 3.02 & 55.3 & 13.92 & -6.4 & 0 \\
& $K=1$ & \underline{0.644} & {\bfseries 0.487} & +0.015 & 3.10 & 48.9 & 13.43 & -9.7 & 0 \\
& $K=2$ & 0.629 & 0.470 & -0.002 & 3.17 & 49.3 & 14.27 & -4.0 & 0 \\
& Full & 0.627 & 0.472 & 0.000 & 2.96 & 49.1 & 14.87 & 0.0 & 0 \\
\midrule
\multirow{4}{*}{CWQ}
& $K=0$ & {\bfseries 0.323} & {\bfseries 0.287} & +0.013 & 7.59 & 147.1 & 36.69 & -12.1 & 0 \\
& $K=1$ & \underline{0.323} & \underline{0.283} & +0.009 & 8.03 & 139.1 & 38.09 & -8.7 & 0 \\
& $K=2$ & 0.314 & 0.277 & +0.003 & 8.57 & 139.1 & 39.98 & -4.2 & 0 \\
& Full & 0.311 & 0.274 & 0.000 & 8.88 & 139.1 & 41.74 & 0.0 & 0 \\
\bottomrule
\end{tabular}
\end{table*}

\begin{figure*}[t]
\centering
\includegraphics[width=.96\linewidth]{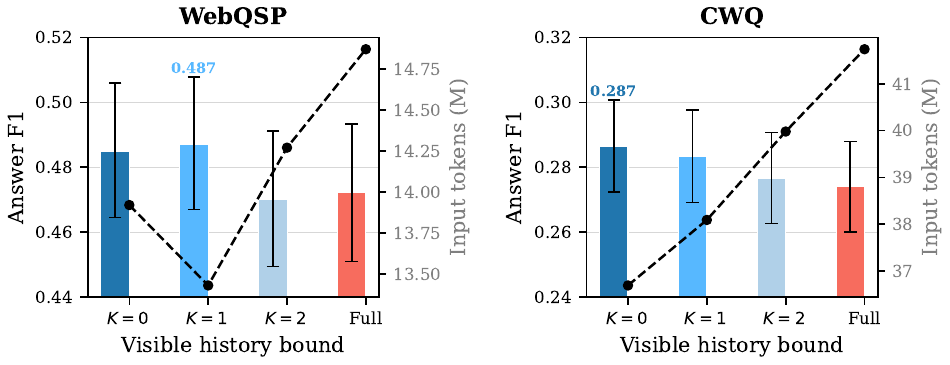}
\caption{Full-test Qwen3.5-9B-AWQ sweep over visible path history. Bars show answer F1 with bootstrap 95\% CI error bars; the dashed line reports estimated input tokens (right axis). Adjacent $K$ values have overlapping CIs; CWQ shows a monotonic decrease as more history is exposed, while WebQSP peaks at one visible hop.}
\label{fig:k-sweep}
\end{figure*}

Table~\ref{tab:qwen35-main} and Figure~\ref{fig:k-sweep} show the central result.
Full-history prompting is not the best setting on either dataset.
On WebQSP, one visible previous hop obtains the highest F1, while $K=0$ gives the highest Hits@1 and is nearly tied in F1.
Both $K=0$ and $K=1$ have higher F1 point estimates than full history.
On CWQ, the most aggressive bound, $K=0$, is best in both Hits@1 and F1.
Increasing visible history from $K=0$ through $K=1$, $K=2$, to full is associated with monotonically decreasing F1 on CWQ.
Adjacent $K$ values have overlapping unpaired bootstrap CIs (e.g., CWQ $K\!=\!0$: $.287\pm.014$, $K\!=\!1$: $.283\pm.014$), so we treat the gaps as favorable point estimates rather than uniformly significant improvements.
Paired bootstrap over per-example F1 gives a positive interval for WebQSP $K\!=\!1$ versus full ($+0.015$, 95\% CI $[+0.003,+0.028]$) and CWQ $K\!=\!0$ versus full ($+0.013$, 95\% CI $[+0.002,+0.024]$), but sign tests over non-tied examples are not significant at $p<.05$.

The WebQSP pattern prevents an over-strong conclusion: no previous hop is not always optimal, since one local hop can help disambiguate the current entity.
In these runs, complete path serialization is not supported as a uniformly better default for each routing call.
The strongest efficiency case is CWQ, where $K=0$ has a higher F1 point estimate while reducing input tokens and wall-clock time.
On WebQSP, bounded contexts have higher F1 point estimates and reduce input tokens, but time is roughly tied because the checkpointed $K=0$ run includes a resume.

\section{Ablations and Controls}
\label{sec:controls}

\begin{table}[t]
\centering
\caption{Model-size and visible-history ablation. Bold marks the best F1 and underlining the second-best within each model--dataset group. The 4B rows show that $K\!=\!1$ is strongest on both datasets; the 9B rows show that bounded histories remain at least competitive with full history.}
\label{tab:model-size-ablation}
\scriptsize
\setlength{\tabcolsep}{4pt}
\begin{tabular}{lll
  S[table-format=1.3]
  S[table-format=1.3]
  S[table-format=1.2]@{h\hspace{.45em}}
  S[table-format=2.2]@{M}}
\toprule
Dataset & Model & History & \multicolumn{1}{c}{Hits@1} & \multicolumn{1}{c}{F1} & \multicolumn{1}{c}{Time} & \multicolumn{1}{c}{In toks} \\
\midrule
WebQSP & 9B & $K=0$ & {\bfseries 0.648} & \underline{0.485} & 3.02 & 13.92 \\
WebQSP & 9B & $K=1$ & \underline{0.644} & {\bfseries 0.487} & 3.10 & 13.43 \\
WebQSP & 9B & Full & 0.627 & 0.472 & 2.96 & 14.87 \\
WebQSP & 4B & $K=0$ & 0.626 & 0.473 & 2.72 & 12.94 \\
WebQSP & 4B & $K=1$ & {\bfseries 0.643} & {\bfseries 0.490} & 2.24 & 11.45 \\
WebQSP & 4B & Full & \underline{0.634} & \underline{0.486} & 2.29 & 12.08 \\
\midrule
CWQ & 9B & $K=0$ & {\bfseries 0.323} & {\bfseries 0.287} & 7.59 & 36.69 \\
CWQ & 9B & $K=1$ & {\bfseries 0.323} & \underline{0.283} & 8.03 & 38.09 \\
CWQ & 9B & Full & 0.311 & 0.274 & 8.88 & 41.74 \\
CWQ & 4B & $K=0$ & 0.294 & 0.265 & 6.66 & 33.39 \\
CWQ & 4B & $K=1$ & {\bfseries 0.321} & {\bfseries 0.297} & 5.66 & 31.58 \\
CWQ & 4B & Full & \underline{0.315} & \underline{0.287} & 3.02 & 17.32 \\
\bottomrule
\end{tabular}
\end{table}

\paragraph{Model size.}
\label{sec:model-size}
Table~\ref{tab:model-size-ablation} shows the most important ablation beyond $K$: model capacity.
At 9B, bounded histories have higher F1 point estimates than full history on both datasets, but the best bound differs: $K=1$ has the highest F1 on WebQSP, while $K=0$ has the highest F1 on CWQ.
At 4B, the picture is more nuanced than a simple reversal of the 9B pattern.
Full history has higher F1 than $K=0$ at 4B (F1 0.486 vs.\ 0.473 on WebQSP, 0.287 vs.\ 0.265 on CWQ), indicating that the smaller model benefits from some visible path context.
However, $K=1$ has higher F1 point estimates than both $K=0$ and full history on both 4B datasets (F1 $0.490\pm0.020$ on WebQSP, $0.297\pm0.014$ on CWQ).
The common pattern across model sizes is that $K=1$ is competitive or has the best point estimate everywhere in our evaluated settings, making it a reasonable candidate default when a single fixed bound is preferred.

\paragraph{Controls.}
The random-relation control checks whether graph access alone explains the result.
It does not: at 9B, random obtains 0.186 F1 on WebQSP and 0.147 on CWQ, compared with 0.485 and 0.287 for BPC $K\!=\!0$.
The CoT diagnostic control, which gives the LLM only the question text without graph access, obtains 0.260 F1 on WebQSP and 0.188 on CWQ\@.
A Think-on-Graph (ToG) reimplementation \citep{sun2024think}, which adds entity pruning and explicit LLM reasoning checks at each hop, reaches 0.465 F1 on WebQSP and 0.281 on CWQ at 9B under the same local model and graph budget.
These controls indicate that the bounded-history sweep is not removing graph access or semantic relation selection; they are not intended as a comprehensive comparison against all KGQA systems.

\section{Error Analysis}
\label{sec:error-analysis}

To understand \emph{when} path context helps or hurts, we compare per-example F1 between each bounded setting and full history.

\paragraph{Win--loss--tie.}
We count examples where a bounded setting scores strictly higher, lower, or equal in F1 compared with full history; paired 9B counts are reported in Appendix Table~\ref{tab:paired-diagnostics}.
Most examples are ties: at 9B, 71--84\% of examples produce identical final F1 regardless of visible history.
This does not prove that routing decisions are identical, but it indicates that the final answer quality is unchanged for most examples under the evaluated bounds.
Among the examples that do differ, $K\!=\!1$ wins more than it loses in every model--dataset combination, making it the only setting with a uniformly positive win--loss balance.
$K\!=\!0$ is net-positive at 9B but net-negative at 4B, consistent with the aggregate F1 pattern in Table~\ref{tab:model-size-ablation}.

\paragraph{Answer cardinality.}
Splitting by gold-answer count reveals an interaction: at 9B, $K\!=\!0$ is better than full history on single-answer questions (mean $\Delta$F1 $+.027$ on WebQSP, $+.020$ on CWQ), while the advantage shrinks or reverses on multi-answer questions ($-.002$ on 9B WebQSP, $-.010$ on 9B CWQ; $-.032$ on 4B WebQSP).
We treat this as an observational hypothesis about when path history helps, not as a causal conclusion.

\paragraph{Routing failures.}
At 9B, $K\!=\!0$ returns ``no relevant path'' on 31 WebQSP and 149 CWQ examples, compared with 51 and 198 for full history, suggesting but not proving that shorter prompts can help focus on the current relation inventory.

\paragraph{Qualitative patterns.}
Qualitatively, $K\!=\!0$ wins often involve full history introducing a plausible but stale earlier entity, while full-history wins often require a previous hop to disambiguate the current entity.
This is consistent with the $K\!=\!1$ pattern: one local hop can help without reintroducing the complete path.

\section{Discussion}

BPC is most appropriate for graph controllers in which exact symbolic state is already available outside the prompt.
In this local Qwen3.5 setup, the question, current entity, and outgoing relation inventory form a competitive compact decision state, while retained complete paths remain available within the fixed answer-extraction budget.
The 4B results suggest one visible hop as a strong candidate bound; shorter prompts reduce input-token load but should not be mistaken for safer or more factual answers.

\section{Conclusion}

Bounded Path Context separates exact symbolic state from model-visible path text.
Full-test Qwen3.5-9B-AWQ results show that small visible histories are competitive with full-history prompting while reducing input-token load.
Diagnostic controls indicate that both graph grounding and learned relation selection matter in this setup; future work should test adaptive $K$ and stronger graph-agent pipelines.

\section*{Limitations}
\label{sec:limitations}

The experiments cover two Freebase-style datasets and deterministic decoding.
The local full-test results use Qwen3.5-AWQ checkpoints; the API rows in the appendix are smaller compatibility checks ($n$=200 per seed) with mixed directions, so they cannot establish cross-backend generalization.
The CoT, random, and ToG rows are diagnostic controls rather than a complete KGQA leaderboard, and we do not include a prompt-compression or path-summary baseline.

The CWQ full-history run uses \texttt{max\_model\_len}=4096 while bounded-context runs use 2048, because some full-history examples exceed the shorter limit.
This difference in vLLM KV-cache allocation is a potential confound for both wall-clock time and accuracy; we cannot fully rule out that part of the full-history degradation on CWQ reflects serving-configuration effects rather than path-content effects alone.

The relation cap, beam budget, depth, and the answer-extraction budget of eight rendered paths are fixed in the reported runs; we do not ablate these choices, and different $K$ values can still change which retained complete paths reach the final answer prompt.
The final result JSONL files do not store the original graph triples, so graph-size and relation-cap coverage statistics are not recoverable from the released result rows alone.
We evaluate fixed $K$ values rather than an adaptive policy, and simple string normalization may miscount correct aliases.
Wall-clock measurements are tied to RTX 2080 Ti hardware and vLLM configuration.
Finally, BPC changes the prompt interface but does not solve incomplete graph neighborhoods, ambiguous topic entities, or answer normalization errors.

\section*{Ethical Considerations}

BPC uses benchmark knowledge graphs and does not introduce new human-subject data.
KGQA systems can reproduce outdated or biased graph facts, and LLM answer extraction can hallucinate or over-select entities.
Deployed systems should expose uncertainty when graph evidence is sparse or ambiguous.
Because BPC keeps retained symbolic paths, it can support audit trails, but those traces should not be mistaken for guarantees of factual correctness.

\paragraph{Licenses.}
WebQSP \citep{yih2016webquestionssp} and CWQ \citep{talmor2018complexwebquestions} are public research benchmarks; we use the RoG-aligned versions hosted on HuggingFace by \citet{luo2024rog} under the MIT license.
Qwen3.5-AWQ checkpoints are distributed under the Apache~2.0 license, and vLLM \citep{kwon2023efficient} is released under Apache~2.0.
Our released code is distributed under the MIT license.
All artifacts are used consistently with their original terms.

\bibliography{custom}

\begin{thebibliography}{37}
\providecommand{\natexlab}[1]{#1}

\bibitem[{Allamanis et~al.(2018)Allamanis, Brockschmidt, and
  Khademi}]{allamanis2018learning}
Miltiadis Allamanis, Marc Brockschmidt, and Mahmoud Khademi. 2018.
\newblock Learning to represent programs with graphs.
\newblock In \emph{Proceedings of the International Conference on Learning
  Representations}.

\bibitem[{Alon et~al.(2019{\natexlab{a}})Alon, Brody, Levy, and
  Yahav}]{alon2019code2seq}
Uri Alon, Shaked Brody, Omer Levy, and Eran Yahav. 2019{\natexlab{a}}.
\newblock code2seq: Generating sequences from structured representations of
  code.
\newblock In \emph{Proceedings of the International Conference on Learning
  Representations}.

\bibitem[{Alon et~al.(2019{\natexlab{b}})Alon, Zilberstein, Levy, and
  Yahav}]{alon2019code2vec}
Uri Alon, Meital Zilberstein, Omer Levy, and Eran Yahav. 2019{\natexlab{b}}.
\newblock code2vec: Learning distributed representations of code.
\newblock \emph{Proceedings of the ACM on Programming Languages},
  3(POPL):1--29.

\bibitem[{Berant et~al.(2013)Berant, Chou, Frostig, and
  Liang}]{berant2013semantic}
Jonathan Berant, Andrew Chou, Roy Frostig, and Percy Liang. 2013.
\newblock Semantic parsing on freebase from question-answer pairs.
\newblock In \emph{Proceedings of the 2013 Conference on Empirical Methods in
  Natural Language Processing}, pages 1533--1544.

\bibitem[{Fang et~al.(2024)Fang, Zhu, and Gurevych}]{fang2024dara}
Haishuo Fang, Xiaodan Zhu, and Iryna Gurevych. 2024.
\newblock {DARA}: Decomposition-alignment-reasoning autonomous language agent
  for question answering over knowledge graphs.
\newblock In \emph{Findings of the Association for Computational Linguistics:
  ACL 2024}, pages 3406--3432.

\bibitem[{Gu et~al.(2023)Gu, Deng, and Su}]{gu2023pangu}
Yu~Gu, Xiang Deng, and Yu~Su. 2023.
\newblock Don't generate, discriminate: A proposal for grounding language
  models to real-world environments.
\newblock In \emph{Proceedings of the 61st Annual Meeting of the Association
  for Computational Linguistics}, pages 4928--4949.

\bibitem[{He et~al.(2021)He, Lan, Jiang, Zhao, and Wen}]{he2021nsm}
Gaole He, Yunshi Lan, Jing Jiang, Wayne~Xin Zhao, and Ji-Rong Wen. 2021.
\newblock Improving multi-hop knowledge base question answering by learning
  intermediate supervision signals.
\newblock In \emph{Proceedings of the 14th ACM International Conference on Web
  Search and Data Mining}, pages 553--561.

\bibitem[{Jiang et~al.(2024)Jiang, Wu, Luo, Li, Lin, Yang, and
  Qiu}]{jiang2023longllmlingua}
Huiqiang Jiang, Qianhui Wu, Xufang Luo, Dongsheng Li, Chin-Yew Lin, Yuqing
  Yang, and Lili Qiu. 2024.
\newblock Longllmlingua: Accelerating and enhancing llms in long context
  scenarios via prompt compression.
\newblock In \emph{Proceedings of the 62nd Annual Meeting of the Association
  for Computational Linguistics}, pages 1658--1677.

\bibitem[{Jiang et~al.(2023{\natexlab{a}})Jiang, Zhou, Dong, Ye, Zhao, and
  Wen}]{jiang2023structgpt}
Jinhao Jiang, Kun Zhou, Zican Dong, Keming Ye, Wayne~Xin Zhao, and Ji-Rong Wen.
  2023{\natexlab{a}}.
\newblock Structgpt: A general framework for large language model to reason
  over structured data.
\newblock In \emph{Proceedings of the 2023 Conference on Empirical Methods in
  Natural Language Processing}, pages 9237--9251.

\bibitem[{Jiang et~al.(2023{\natexlab{b}})Jiang, Zhou, Zhao, and
  Wen}]{jiang2023unikgqa}
Jinhao Jiang, Kun Zhou, Wayne~Xin Zhao, and Ji-Rong Wen. 2023{\natexlab{b}}.
\newblock Unikgqa: Unified retrieval and reasoning for solving multi-hop
  question answering over knowledge graph.
\newblock In \emph{Proceedings of the International Conference on Learning
  Representations}.

\bibitem[{Kwon et~al.(2023)Kwon, Li, Zhuang, Sheng, Zheng, Yu, Gonzalez, Zhang,
  and Stoica}]{kwon2023efficient}
Woosuk Kwon, Zhuohan Li, Siyuan Zhuang, Ying Sheng, Lianmin Zheng, Cody~Hao Yu,
  Joseph~E. Gonzalez, Hao Zhang, and Ion Stoica. 2023.
\newblock Efficient memory management for large language model serving with
  {PagedAttention}.
\newblock In \emph{Proceedings of the 29th Symposium on Operating Systems
  Principles}.

\bibitem[{Lan and Jiang(2020)}]{lan2020query}
Yunshi Lan and Jing Jiang. 2020.
\newblock Query graph generation for answering multi-hop complex questions from
  knowledge bases.
\newblock In \emph{Proceedings of the 58th Annual Meeting of the Association
  for Computational Linguistics}, pages 969--974.

\bibitem[{Li et~al.(2025{\natexlab{a}})Li, Miao, and Li}]{li2025subgraphrag}
Mufei Li, Siqi Miao, and Pan Li. 2025{\natexlab{a}}.
\newblock Simple is effective: The roles of graphs and large language models in
  knowledge-graph-based retrieval-augmented generation.
\newblock In \emph{Proceedings of the International Conference on Learning
  Representations}.

\bibitem[{Li et~al.(2023{\natexlab{a}})Li, Ma, Zhuang, Gu, Su, and
  Chen}]{li2023kbbinder}
Tianle Li, Xueguang Ma, Alex Zhuang, Yu~Gu, Yu~Su, and Wenhu Chen.
  2023{\natexlab{a}}.
\newblock Few-shot in-context learning for knowledge base question answering.
\newblock In \emph{Proceedings of the 61st Annual Meeting of the Association
  for Computational Linguistics}, pages 6966--6981.

\bibitem[{Li et~al.(2024)Li, Song, Zhou, Tian, Wang, Yang, and
  Zhang}]{li2024keldar}
Yading Li, Dandan Song, Changzhi Zhou, Yuhang Tian, Hao Wang, Ziyi Yang, and
  Shuhao Zhang. 2024.
\newblock A framework of knowledge graph-enhanced large language model based on
  question decomposition and atomic retrieval.
\newblock In \emph{Findings of the Association for Computational Linguistics:
  EMNLP 2024}, pages 11472--11485.

\bibitem[{Li et~al.(2023{\natexlab{b}})Li, Dong, Guerin, and
  Lin}]{li2023selectivecontext}
Yucheng Li, Bo~Dong, Frank Guerin, and Chenghua Lin. 2023{\natexlab{b}}.
\newblock Compressing context to enhance inference efficiency of large language
  models.
\newblock In \emph{Proceedings of the 2023 Conference on Empirical Methods in
  Natural Language Processing}, pages 6342--6353.

\bibitem[{Li et~al.(2025{\natexlab{b}})Li, Liu, Su, and
  Collier}]{li2025promptcompression}
Zongqian Li, Yinhong Liu, Yixuan Su, and Nigel Collier. 2025{\natexlab{b}}.
\newblock Prompt compression for large language models: A survey.
\newblock In \emph{Proceedings of the 2025 Conference of the Nations of the
  Americas Chapter of the Association for Computational Linguistics: Human
  Language Technologies (Volume 1: Long Papers)}, pages 7182--7195.

\bibitem[{Liu et~al.(2024)Liu, Lin, Hewitt, Paranjape, Bevilacqua, Petroni, and
  Liang}]{liu2024lost}
Nelson~F. Liu, Kevin Lin, John Hewitt, Ashwin Paranjape, Michele Bevilacqua,
  Fabio Petroni, and Percy Liang. 2024.
\newblock Lost in the middle: How language models use long contexts.
\newblock \emph{Transactions of the Association for Computational Linguistics},
  12:157--173.

\bibitem[{Luo et~al.(2024{\natexlab{a}})Luo, E, Tang, Peng, Guo, Zhang, Ma,
  Dong, Song, Lin, Zhu, and Luu}]{luo2024chatkbqa}
Haoran Luo, Haihong E, Zichen Tang, Shiyao Peng, Yikai Guo, Wentai Zhang,
  Chenghao Ma, Guanting Dong, Meina Song, Wei Lin, Yifan Zhu, and Anh~Tuan Luu.
  2024{\natexlab{a}}.
\newblock {ChatKBQA}: A generate-then-retrieve framework for knowledge base
  question answering with fine-tuned large language models.
\newblock In \emph{Findings of the Association for Computational Linguistics:
  ACL 2024}, pages 2039--2056.

\bibitem[{Luo et~al.(2024{\natexlab{b}})Luo, Li, Haffari, and Pan}]{luo2024rog}
Linhao Luo, Yuan-Fang Li, Gholamreza Haffari, and Shirui Pan.
  2024{\natexlab{b}}.
\newblock Reasoning on graphs: Faithful and interpretable large language model
  reasoning.
\newblock In \emph{Proceedings of the International Conference on Learning
  Representations}.

\bibitem[{Luo et~al.(2025)Luo, Zhao, Haffari, Li, Gong, and Pan}]{luo2024gcr}
Linhao Luo, Zicheng Zhao, Gholamreza Haffari, Yuan-Fang Li, Chen Gong, and
  Shirui Pan. 2025.
\newblock Graph-constrained reasoning: Faithful reasoning on knowledge graphs
  with large language models.
\newblock In \emph{Proceedings of the 42nd International Conference on Machine
  Learning}, volume 267.

\bibitem[{Mavromatis and Karypis(2025)}]{mavromatis2024gnnrag}
Costas Mavromatis and George Karypis. 2025.
\newblock {GNN}-{RAG}: Graph neural retrieval for efficient large language
  model reasoning on knowledge graphs.
\newblock In \emph{Findings of the Association for Computational Linguistics:
  ACL 2025}, pages 16682--16699.

\bibitem[{Pan et~al.(2024)Pan, Wu, Jiang, Xia, Luo, Zhang, Lin, R{\"u}hle,
  Yang, Qiu, and Zhang}]{pan2024llmlingua2}
Zhuoshi Pan, Qianhui Wu, Huiqiang Jiang, Mengzhou Xia, Xufang Luo, Jue Zhang,
  Qingwei Lin, Victor R{\"u}hle, Yuqing Yang, Lili Qiu, and Dongmei Zhang.
  2024.
\newblock Llmlingua-2: Data distillation for efficient and faithful
  task-agnostic prompt compression.
\newblock In \emph{Findings of the Association for Computational Linguistics:
  ACL 2024}, pages 963--981.

\bibitem[{{Qwen}(2024)}]{qwen2025qwen25}
{Qwen}. 2024.
\newblock {Qwen2.5} technical report.
\newblock \emph{arXiv preprint arXiv:2412.15115}.

\bibitem[{Saxena et~al.(2020)Saxena, Tripathi, and
  Talukdar}]{saxena2020embedkgqa}
Apoorv Saxena, Aditay Tripathi, and Partha Talukdar. 2020.
\newblock Improving multi-hop question answering over knowledge graphs using
  knowledge base embeddings.
\newblock In \emph{Proceedings of the 58th Annual Meeting of the Association
  for Computational Linguistics}, pages 4498--4507.

\bibitem[{Shi et~al.(2021)Shi, Cao, Hou, Li, and Zhang}]{shi2021transfernet}
Jiaxin Shi, Shulin Cao, Lei Hou, Juanzi Li, and Hanwang Zhang. 2021.
\newblock {TransferNet}: An effective and transparent framework for multi-hop
  question answering over relation graph.
\newblock In \emph{Proceedings of the 2021 Conference on Empirical Methods in
  Natural Language Processing}.

\bibitem[{Sun et~al.(2019)Sun, Bedrax-Weiss, and Cohen}]{sun2019pullnet}
Haitian Sun, Tania Bedrax-Weiss, and William~W. Cohen. 2019.
\newblock {PullNet}: Open domain question answering with iterative retrieval on
  knowledge bases and text.
\newblock In \emph{Proceedings of the 2019 Conference on Empirical Methods in
  Natural Language Processing and the 9th International Joint Conference on
  Natural Language Processing}.

\bibitem[{Sun et~al.(2018)Sun, Dhingra, Zaheer, Mazaitis, Salakhutdinov, and
  Cohen}]{sun2018graftnet}
Haitian Sun, Bhuwan Dhingra, Manzil Zaheer, Kathryn Mazaitis, Ruslan
  Salakhutdinov, and William~W. Cohen. 2018.
\newblock Open domain question answering using early fusion of knowledge bases
  and text.
\newblock In \emph{Proceedings of the 2018 Conference on Empirical Methods in
  Natural Language Processing}, pages 4231--4242.

\bibitem[{Sun et~al.(2024)Sun, Xu, Tang, Wang, Lin, Gong, Ni, Shum, Guo, and
  Zhang}]{sun2024think}
Jiashuo Sun, Chengjin Xu, Luming Tang, Saizhuo Wang, Chen Lin, Yeyun Gong,
  Lionel~M. Ni, Heung-Yeung Shum, Jian Guo, and Nan Zhang. 2024.
\newblock Think-on-graph: Deep and responsible reasoning of large language
  model on knowledge graph.
\newblock In \emph{Proceedings of the International Conference on Learning
  Representations}.

\bibitem[{Talmor and Berant(2018)}]{talmor2018complexwebquestions}
Alon Talmor and Jonathan Berant. 2018.
\newblock The web as a knowledge-base for answering complex questions.
\newblock In \emph{Proceedings of the 2018 Conference of the North American
  Chapter of the Association for Computational Linguistics}, pages 641--651.

\bibitem[{Wei et~al.(2022)Wei, Wang, Schuurmans, Bosma, Ichter, Xia, Chi, Le,
  and Zhou}]{wei2022chain}
Jason Wei, Xuezhi Wang, Dale Schuurmans, Maarten Bosma, Brian Ichter, Fei Xia,
  Ed~Chi, Quoc Le, and Denny Zhou. 2022.
\newblock Chain-of-thought prompting elicits reasoning in large language
  models.
\newblock \emph{Advances in Neural Information Processing Systems},
  35:24824--24837.

\bibitem[{Xu et~al.(2024)Xu, Shi, and Choi}]{xu2024recomp}
Fangyuan Xu, Weijia Shi, and Eunsol Choi. 2024.
\newblock Recomp: Improving retrieval-augmented lms with context compression
  and selective augmentation.
\newblock In \emph{Proceedings of the International Conference on Learning
  Representations}.

\bibitem[{Yao et~al.(2023)Yao, Zhao, Yu, Du, Shafran, Narasimhan, and
  Cao}]{yao2023react}
Shunyu Yao, Jeffrey Zhao, Dian Yu, Nan Du, Izhak Shafran, Karthik Narasimhan,
  and Yuan Cao. 2023.
\newblock React: Synergizing reasoning and acting in language models.
\newblock In \emph{Proceedings of the International Conference on Learning
  Representations}.

\bibitem[{Yasunaga et~al.(2021)Yasunaga, Ren, Bosselut, Liang, and
  Leskovec}]{yasunaga2021qagnn}
Michihiro Yasunaga, Hongyu Ren, Antoine Bosselut, Percy Liang, and Jure
  Leskovec. 2021.
\newblock {QA-GNN}: Reasoning with language models and knowledge graphs for
  question answering.
\newblock In \emph{Proceedings of the 2021 Conference of the North American
  Chapter of the Association for Computational Linguistics}.

\bibitem[{Yih et~al.(2016)Yih, Richardson, Meek, Chang, and
  Suh}]{yih2016webquestionssp}
Wen-tau Yih, Matthew Richardson, Christopher Meek, Ming-Wei Chang, and Jina
  Suh. 2016.
\newblock A value-based search method for knowledge base question answering.
\newblock In \emph{Proceedings of the 54th Annual Meeting of the Association
  for Computational Linguistics}, pages 505--515.

\bibitem[{Yu et~al.(2023)Yu, Zhang, Ng, Zhu, Li, Wang, Hu, Wang, Wang, and
  Hakkani-Tur}]{yu2023decaf}
Donghan Yu, Sheng Zhang, Patrick Ng, Henghui Zhu, Alexander~Hanbo Li, Jun Wang,
  Yiqun Hu, William Wang, Zhiguo Wang, and Dilek Hakkani-Tur. 2023.
\newblock Decaf: Joint decoding of answers and logical forms for knowledge base
  question answering.
\newblock In \emph{Proceedings of the International Conference on Learning
  Representations}.

\bibitem[{Zhang et~al.(2022)Zhang, Zhang, Yu, Tang, Tang, Li, and
  Chen}]{zhang2022subgraph}
Jing Zhang, Xiaokang Zhang, Jifan Yu, Jian Tang, Jie Tang, Cuiping Li, and Hong
  Chen. 2022.
\newblock Subgraph retrieval enhanced model for multi-hop knowledge base
  question answering.
\newblock In \emph{Proceedings of the 60th Annual Meeting of the Association
  for Computational Linguistics}, pages 5773--5784.

\end{thebibliography}

\appendix

\section{Prompt Interface}

The relation-selection prompt has four logical fields: the question, the visible path history, the current entity, and candidate outgoing relations.
For $K=0$, the history field is rendered as an explicit no-history marker rather than removed, so formatting remains stable across settings.
For $K=1$ and $K=2$, the field contains only the most recent one or two hops.
For $K=\mathrm{full}$, it contains the entire retained path.
The final answer prompt receives retained paths after search and is not directly truncated by the relation-selection bound, although the implementation renders only the first eight retained complete paths under a shared prompt budget.

\paragraph{Relation-selection fields.}
The implementation renders relation candidates in a stable order after graph-side filtering.
The prompt asks the model to return a compact relation selection rather than a free-form explanation.
This is important for no-thinking Qwen3.5 inference: short deterministic outputs reduce the chance that the model spends tokens on rationale text rather than the controller action.

\paragraph{Answer extraction fields.}
The answer extractor receives the question and a fixed-budget rendering of retained symbolic paths.
This is where full path strings remain useful.
BPC therefore delays complete path serialization until the stage where path evidence is directly relevant, while preserving retained path objects for audit outside the routing prompt.
This distinction is the core reason BPC can reduce routing context without removing retained evidence paths from the controller.

\section{API Compatibility Checks}

\begin{table}[ht]
\centering
\caption{Small API compatibility checks ($n$=200 per seed). These rows are mixed: seed 123 favors $K=0$ on both datasets, while seed 42 is close and does not consistently favor $K=0$ in F1. We therefore exclude these rows from the main claim.}
\label{tab:app-deepseek}
\scriptsize
\setlength{\tabcolsep}{4pt}
\begin{tabular}{lrrrrr}
\toprule
Dataset & Seed & \multicolumn{1}{c}{$K=0$ H@1} & \multicolumn{1}{c}{$K=0$ F1} & \multicolumn{1}{c}{Full H@1} & \multicolumn{1}{c}{Full F1} \\
\midrule
WebQSP & 42  & 0.625 & 0.474 & 0.610 & 0.501 \\
WebQSP & 123 & 0.705 & 0.508 & 0.610 & 0.473 \\
CWQ    & 42  & 0.345 & 0.298 & 0.340 & 0.299 \\
CWQ    & 123 & 0.420 & 0.397 & 0.370 & 0.364 \\
\bottomrule
\end{tabular}
\end{table}

Table~\ref{tab:app-deepseek} gives the API compatibility rows available in the current result folder.
The runs are deliberately separated from the local Qwen3.5 manifest.
They use the same BPC interface and decoding temperature, but API serving behavior and sample composition are not controlled as tightly as the full local runs.
Because the direction varies by seed, we treat these rows as a reminder that backend transfer remains open rather than as supporting evidence for the main claim.

\section{Paired and Routing Diagnostics}

\begin{table}[ht]
\centering
\caption{Paired tests against full history for the 9B main sweep. CIs are paired bootstrap intervals over per-example F1 differences. The sign test uses only non-tied examples, so it is conservative when most examples tie.}
\label{tab:paired-diagnostics}
\scriptsize
\setlength{\tabcolsep}{4pt}
\begin{tabular}{llrrrr}
\toprule
Dataset & vs Full & $\Delta$F1 & 95\% CI & W/L/T & Sign $p$ \\
\midrule
WebQSP & $K=0$ & +.013 & [-.003,.028] & 247/218/1163 & .194 \\
WebQSP & $K=1$ & +.015 & [+.003,+.028] & 192/162/1274 & .123 \\
WebQSP & $K=2$ & -.002 & [-.013,+.009] & 132/132/1364 & 1.000 \\
CWQ & $K=0$ & +.013 & [+.002,+.024] & 297/262/2972 & .150 \\
CWQ & $K=1$ & +.009 & [+.000,+.019] & 235/196/3100 & .067 \\
CWQ & $K=2$ & +.003 & [-.005,+.010] & 152/151/3228 & 1.000 \\
\bottomrule
\end{tabular}
\end{table}

\begin{table}[ht]
\centering
\caption{Routing-depth diagnostics recoverable from the final JSONL rows. The result files store search summaries but not original graph triples, so graph-size and relation-cap coverage statistics require the original dataset files.}
\label{tab:routing-depth}
\scriptsize
\setlength{\tabcolsep}{4pt}
\begin{tabular}{llrrr}
\toprule
Dataset & History & Avg. depth & Depth 5 & No-answer \\
\midrule
WebQSP & $K=0$ & 4.79 & 93.2\% & 39 \\
WebQSP & $K=1$ & 4.55 & 85.9\% & 43 \\
WebQSP & $K=2$ & 4.44 & 82.4\% & 49 \\
WebQSP & Full & 4.42 & 80.7\% & 54 \\
CWQ & $K=0$ & 4.93 & 97.9\% & 167 \\
CWQ & $K=1$ & 4.86 & 95.0\% & 202 \\
CWQ & $K=2$ & 4.80 & 93.1\% & 212 \\
CWQ & Full & 4.78 & 91.7\% & 202 \\
\bottomrule
\end{tabular}
\end{table}

\section{Result Manifest}

\begin{table*}[t]
\centering
\caption{Reported local Qwen3.5 result manifest. Bold and underlining mark the best and second-best F1 within each model--dataset block. Bounded BPC rows are compared against full-history prompting; random, CoT, and ToG rows are diagnostic controls.}
\label{tab:manifest}
\scriptsize
\setlength{\tabcolsep}{3pt}
\begin{tabular}{lllrrrrrr}
\toprule
Model & Dataset & Method/$K$ & H@1 & F1 & Time & Calls & In toks & Out toks \\
\midrule
9B & WebQSP & BPC/0 & \textbf{0.648} & \underline{0.485} & 3.02h & 55.3K & 13.92M & 0.75M \\
9B & WebQSP & BPC/1 & \underline{0.644} & \textbf{0.487} & 3.10h & 48.9K & 13.43M & 0.60M \\
9B & WebQSP & BPC/2 & 0.629 & 0.470 & 3.17h & 49.3K & 14.27M & 0.68M \\
9B & WebQSP & BPC/full & 0.627 & 0.472 & 2.96h & 49.1K & 14.87M & 0.67M \\
9B & WebQSP & CoT & 0.340 & 0.260 & 0.05h & 1.6K & 0.09M & 0.02M \\
9B & WebQSP & Random/0 & 0.244 & 0.186 & 0.25h & 1.6K & 1.18M & 0.01M \\
9B & WebQSP & ToG & 0.568 & 0.465 & 2.19h & 25.1K & 5.84M & 0.37M \\
9B & CWQ & BPC/0 & \textbf{0.323} & \textbf{0.287} & 7.59h & 147.1K & 36.69M & 1.99M \\
9B & CWQ & BPC/1 & \textbf{0.323} & \underline{0.283} & 8.03h & 139.1K & 38.09M & 1.78M \\
9B & CWQ & BPC/2 & 0.314 & 0.277 & 8.57h & 139.1K & 39.98M & 1.95M \\
9B & CWQ & BPC/full & 0.311 & 0.274 & 8.88h & 139.1K & 41.74M & 1.91M \\
9B & CWQ & CoT & 0.202 & 0.188 & 0.08h & 3.5K & 0.23M & 0.02M \\
9B & CWQ & Random/0 & 0.159 & 0.147 & 0.54h & 3.5K & 2.52M & 0.02M \\
9B & CWQ & ToG & 0.304 & 0.281 & 8.55h & 124.1K & 28.95M & 1.89M \\
\midrule
4B & WebQSP & BPC/0 & 0.626 & 0.473 & 2.72h & 49.7K & 12.94M & 0.58M \\
4B & WebQSP & BPC/1 & \textbf{0.643} & \textbf{0.490} & 2.24h & 39.5K & 11.45M & 0.35M \\
4B & WebQSP & BPC/2 & 0.628 & 0.481 & 2.63h & 38.5K & 11.59M & 0.36M \\
4B & WebQSP & BPC/full & \underline{0.634} & \underline{0.486} & 2.29h & 38.6K & 12.08M & 0.37M \\
4B & WebQSP & CoT & 0.275 & 0.202 & 0.07h & 1.6K & 0.09M & 0.01M \\
4B & WebQSP & Random/0 & 0.216 & 0.164 & 0.19h & 1.6K & 1.18M & 0.01M \\
4B & WebQSP & ToG & 0.616 & 0.498 & 2.55h & 40.0K & 9.50M & 0.67M \\
4B & CWQ & BPC/0 & 0.294 & 0.265 & 6.66h & 131.9K & 33.39M & 1.53M \\
4B & CWQ & BPC/1 & \textbf{0.321} & \textbf{0.297} & 5.66h & 111.9K & 31.58M & 0.92M \\
4B & CWQ & BPC/2 & 0.312 & 0.286 & 5.94h & 99.1K & 29.32M & 0.84M \\
4B & CWQ & BPC/full & \underline{0.315} & \underline{0.287} & 3.02h & 56.8K & 17.32M & 0.50M \\
4B & CWQ & CoT & 0.170 & 0.155 & 0.13h & 3.5K & 0.23M & 0.02M \\
4B & CWQ & Random/0 & 0.127 & 0.117 & 0.38h & 3.5K & 2.52M & 0.01M \\
4B & CWQ & ToG & 0.259 & 0.238 & 9.77h & 186.3K & 44.67M & 3.17M \\
\bottomrule
\end{tabular}
\end{table*}

Table~\ref{tab:manifest} gives the reported local result inventory, including CoT, random, and ToG diagnostic controls.
The 9B rows support the central BPC claim: bounded histories are competitive with full-history prompting in this setting while reducing input tokens.
The 4B rows show a model-size interaction discussed in Section~\ref{sec:model-size}: while $K\!=\!0$ is weaker than full history at 4B, $K\!=\!1$ has the strongest 4B point estimates and is competitive at 9B, suggesting that one hop of visible history is a strong candidate fixed bound in the evaluated settings.

\begin{table*}[t]
\centering
\caption{Provenance for the 9B main rows. Paths are relative to the released result directory. The WebQSP $K=0$ cost uses reconstructed accounting from checkpointed segments; accuracy uses the final complete JSONL.}
\label{tab:provenance}
\scriptsize
\setlength{\tabcolsep}{3pt}
\begin{tabular}{lll p{.42\linewidth} l l}
\toprule
Dataset & History & Ctx. & JSONL directory & TP & Accounting note \\
\midrule
WebQSP & $K=0$ & 2048 & \texttt{run\_9b\_bpc\_k0} & 2 & reconstructed cost \\
WebQSP & $K=1$ & 2048 & \texttt{run\_9b\_bpc\_k1\_k2} & 2 & direct summary \\
WebQSP & $K=2$ & 2048 & \texttt{run\_9b\_bpc\_k1\_k2} & 2 & direct summary \\
WebQSP & Full & 2048 & \texttt{run\_9b\_bpc\_kfull\_webqsp} & 2 & direct summary \\
CWQ & $K=0$ & 2048 & \texttt{run\_9b\_bpc\_k0} & 2 & direct summary \\
CWQ & $K=1$ & 2048 & \texttt{run\_9b\_bpc\_k1\_k2} & 2 & direct summary \\
CWQ & $K=2$ & 2048 & \texttt{run\_9b\_bpc\_k1\_k2} & 2 & direct summary \\
CWQ & Full & 4096 & \texttt{run\_9b\_bpc\_kfull\_cwq} & 2 & different context length \\
\bottomrule
\end{tabular}
\end{table*}

\section{Reporting Scope}

The paper's main empirical scope is intentionally narrow.
We focus on Qwen3.5 local inference; the small API checks in Table~\ref{tab:app-deepseek} are reported only for transparency and are not used as primary evidence.
Failed full-precision attempts, OOM retries, smoke tests, and superseded partial runs are archived but not averaged into the reported rows.
The manifest in Table~\ref{tab:manifest} gives reviewers a compact audit trail for the rows used in the paper, while the result directory retains superseded artifacts separately.

\section{Discarded and Superseded Artifacts}

The local result directory contains archived intermediate packages from earlier development.
The paper tables use the final Qwen3.5 summaries and the paper-specific manifest.
Superseded runs include smoke tests, failed full-precision attempts that did not fit on the available GPUs, and interrupted retries whose completed rows were later superseded by complete checkpoint-resumed runs.
These artifacts are retained for auditability but are not mixed into the reported aggregates.
This policy avoids a common failure mode in long-running local experiments: accidentally averaging a failed retry with a completed run.

\section{Reproducibility Notes}

All local runs use deterministic decoding with temperature 0.
The main controller uses beam width 3, maximum depth 5, relation cap 50, and a maximum of 16 retained beams.
For long local runs, checkpoints are written after each example.
If a run is resumed, already completed rows are skipped.
The result package therefore contains both final summaries and, where relevant, log files documenting restarts.
The paper table uses final complete JSONL files for accuracy, with corrected accounting for checkpointed runs.

\end{document}